\documentclass[a4paper,twoside]{article}

\usepackage{epsfig}
\usepackage{subcaption}
\usepackage{calc}
\usepackage{amstext}
\usepackage{amsthm}
\usepackage{multicol}
\usepackage{pslatex}
\usepackage{apalike}
\usepackage{cite}
\usepackage{amsmath,amssymb,amsfonts}
\usepackage{algorithmic}
\usepackage{graphicx}
\usepackage{textcomp}
\usepackage{xcolor}
\usepackage{multirow}
\usepackage[linesnumbered,ruled]{algorithm2e}

\usepackage{booktabs}
\usepackage{hyperref}

\usepackage{SCITEPRESS}     

\begin{document}

\title{Unsupervised Gaze Prediction in Egocentric Videos by Energy-based Surprise Modeling}

\author{\authorname{Sathyanarayanan N. Aakur\sup{1} and Arunkumar Bagavathi\sup{1}}
\affiliation{\sup{1}Department of Computer Science, Oklahoma State University, Stillwater, OK, USA 74078}
\email{\{saakurn, abagava\}@okstate.edu}
}

\abstract{
Egocentric perception has grown rapidly with the advent of immersive computing devices. Human gaze prediction is an important problem in analyzing egocentric videos and has primarily been tackled through either saliency-based modeling or highly supervised learning. We quantitatively analyze the generalization capabilities of supervised, deep learning models on the egocentric gaze prediction task on unseen, out-of-domain data. We find that their performance is highly dependent on the training data and is restricted to the domains specified in the training annotations. 
In this work, we tackle the problem of jointly predicting human gaze points and temporal segmentation of egocentric videos without using any training data. 
We introduce an \textit{unsupervised} computational model that draws inspiration from cognitive psychology models of event perception. 
We use Grenander's pattern theory formalism to represent spatial-temporal features and model \textit{surprise} as a mechanism to predict gaze fixation points. Extensive evaluation on two publicly available datasets - GTEA and GTEA+ datasets-shows that the proposed model can significantly outperform all unsupervised baselines and some supervised gaze prediction baselines. Finally, we show that the model can also temporally segment egocentric videos with a performance comparable to more complex, fully supervised deep learning baselines.
}

\onecolumn \maketitle \normalsize \setcounter{footnote}{0} \vfill

\section{Introduction}\label{sec:intro}
\begin{figure*}[ht]
\centering
\includegraphics[width=0.95\textwidth]{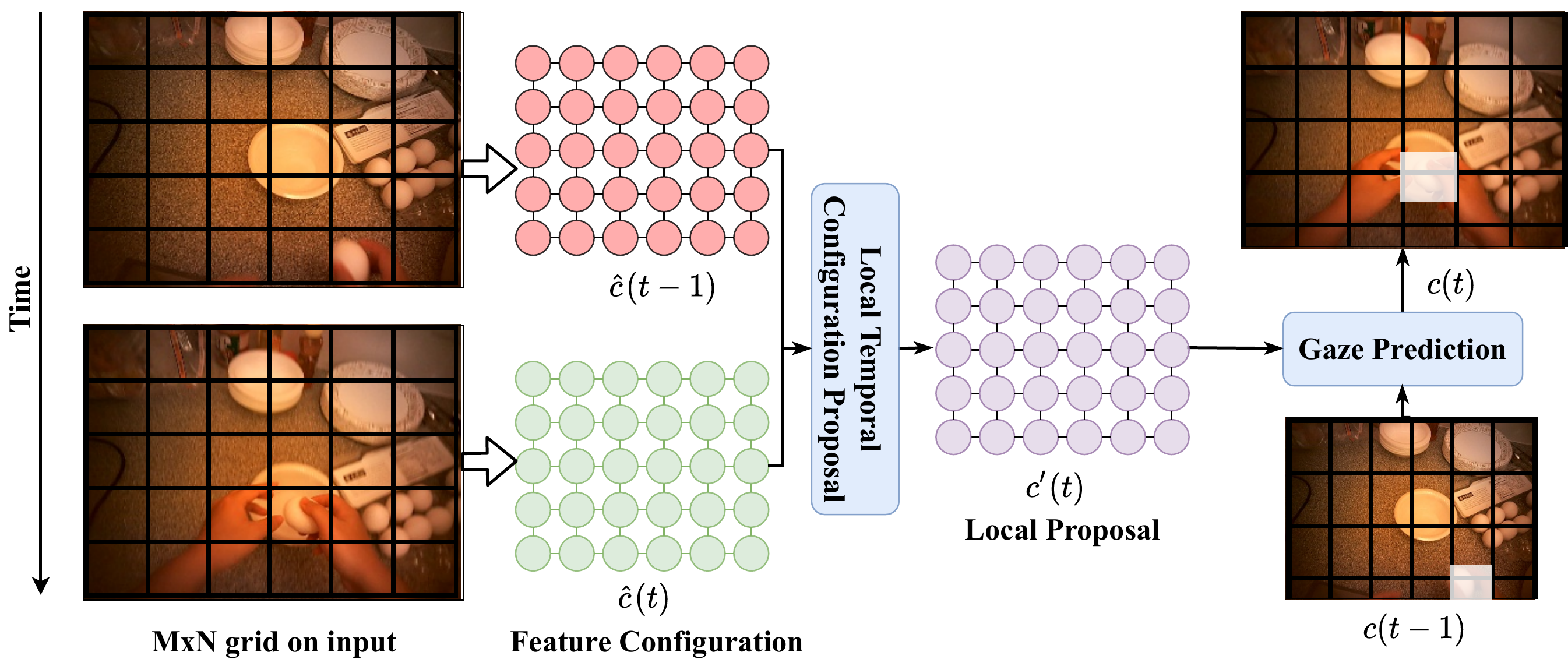}
\caption{\textbf{Overall Approach}: The proposed approach consists of three essential stages: constructing the feature configuration, a local temporal configuration proposal and the final gaze prediction. Note that the output of the gaze prediction module is another configuration which is visualized as an attention map.
} 
\label{fig:overallArch}
\end{figure*}
The emergence of wearable and immersive computing devices for virtual and augmented reality has enabled the acquisition of images and videos from a first-person perspective. Given the recent advances in computer vision, egocentric analysis could be used to infer the surrounding scene and enhance the quality of living through immersive, user-centric applications.
At the core of such an application is the need to \textit{understand} the user's actions and where they are looking. More specifically, gaze prediction is an essential task in egocentric perception. It refers to the process of predicting human fixation points in the scene with respect to the head-centered coordinate system. Beyond enabling more efficient video analysis, studies from psychology have shown that human gaze estimation capture human intent to help collaboration~\cite{huang2015using}. 
While the tasks of activity recognition and segmentation have been explored in recent literature~\cite{lea2016temporal}, we aim to tackle the task of \textit{unsupervised} gaze prediction and temporal event segmentation in a unified framework.

Drawing inspiration from psychology \cite{horstmann2015surprise,horstmann2016novelty,zacks2001perceiving}, we identify that the notion of predictability and \textit{surprise} is a common mechanism to jointly predict gaze and perform temporal segmentation. Defined broadly as the \textit{surprise-attention} hypothesis, studies have found that any deviations from expectations have a strong effect on both attention processing and event perception in humans. Specifically, short-term spatial surprise, such as between subsequent frames of a video, has a high probability of human fixation and affects saccade patterns. Long-term temporal surprise leads to the perception of new events~\cite{zacks2001perceiving}. We leverage such findings and formulate a framework that jointly models both short-term and long-term surprise using Grenander's pattern theory~\cite{grenander1996elements} representations.

A significant departure from recent pattern theory formulations~\cite{aakur2019generating}, our framework models bottom-up, feature-level spatial-temporal correlations in egocentric videos. 
We represent the spatial-temporal structure of egocentric videos using Grenander's pattern theory formalism~\cite{grenander1996elements}. 
The spatial features are encoded in a local \textit{configuration}, whose energy represents the expectation of the model with respect to the recent past. Configurations are aggregated across time to provide a local temporal proposal for capturing the spatial-temporal correlation of video features. An acceptor function is used to switch between saccade and fixation modes to select the final configuration proposal. Localizing the source of maximum surprise provides attention points, and monitoring global surprise allows for temporally segmenting videos.

\textbf{Contributions}: Our contributions are four-fold: (i) We evaluate the ability of supervised gaze prediction models to generalize to out of domain data, (ii) we introduce an unsupervised gaze prediction framework based on surprise modeling that enables gaze prediction \textit{without training} and outperforms all unsupervised baselines and some supervised baselines, (iii) we demonstrate that the pattern theory representation can be used to tackle different tasks such as unsupervised video event segmentation to achieve comparable performance to state-of-the-art deep learning approaches, and (iv) we show that pattern theory representations can be extended to beyond semantic, symbolic reasoning mechanisms.

\section{Related Work}
\textit{Saliency-based models} treat the gaze prediction problem by modeling the visual saliency of a scene and identifying areas that can attract the gaze of a person. At the core of traditional saliency prediction methods is the idea of feature integration~\cite{treisman1980feature}, which is based on the combination of multiple levels of features. Introduced by \textit{Itti et al}~\cite{itti2000saliency,itti2006bayesian}, there have been many approaches to saliency prediction~\cite{bruce2006saliency,leboran2016dynamic,hou2011image,leboran2016dynamic}, including graph-based models~\cite{harel2007graph}, supervised CNN-based saliency~\cite{jiang2015salicon} and video-based saliency~\cite{hossein2015many,leboran2016dynamic}. 

\textit{Supervised Gaze Prediction} has been an increasingly popular way to tackle the problem of gaze prediction in egocentric videos. \textit{Li et al. }~\cite{li2013learning} proposed a graphical model to combine egocentric cues such as camera motion, hand positions, and motion and modeled gaze prediction as a function of these latent variables. 
Deep learning-based approaches have been the other primary modeling option. 
\textit{Zhang et al.}~\cite{zhang2017deep} used a Generative Adversarial Network (GAN) to handle the problem of gaze anticipation. The GAN was used to generate future frames, and a 3D CNN temporal saliency model was used to predict gaze positions. \textit{Huang et al. }~\cite{huang2018predicting} used recurrent neural networks to predict gaze positions by combining task-specific cues with bottom-up saliency. 
While very useful, such deep learning models are increasingly complex, and the amount of training data required by such approaches is enormous. 
\section{Motivation: Why Unsupervised Gaze Prediction?}\label{sec:motivation}
In this section, we analyze the ability of current supervised models to generalize beyond their training domain. 
Most success in egocentric gaze prediction has been through supervised learning. These models~\cite{huang2018predicting,li2013learning,zhang2017deep}, primarily based on deep learning approaches, require large amounts of annotated training data in the form of gaze locations and other auxiliary data such as task label~\cite{huang2018predicting}, camera motion, and hand location~\cite{li2013learning}, to name a few. 
Such information requires manual human annotation and can be expensive to obtain. Additionally, it may not be possible to get \textit{annotated} data for \textit{every eventuality and domains} for training supervised models. 
The combination of these two issues means that current systems are restricted to specific environments. 
We evaluate cross-domain gaze prediction by training supervised models on the GTEA Gaze+ dataset~\cite{li2013learning} and evaluating on the GTEA Gaze dataset~\cite{fathi2012learning}.
\begin{table}
\centering
\resizebox{0.99\columnwidth}{!}{
\begin{tabular}{|c|c|c|c|}
\hline
\multirow{3}{*}{\textbf{Approach}}      & \textbf{Same}  & \textbf{Different} & \multirow{3}{*}{\textbf{$\Delta$}}\\ 
& \textbf{Domain} & \textbf{Domain} & \\
& \textbf{(AAE)} & \textbf{(AAE)} & \\
\toprule
  CNN-LSTM Predictor & $9.82$ & $16.49$ & $\mathbf{6.67}$ \\
  2-Stream CNN & $5.52$ & $11.35$ & $5.83$\\
  2-Stream CNN + LSTM & $5.35$ & $10.71$ & $5.36$\\
  \midrule
  Ours (Unsupervised)* & $\mathbf{11.6}$ & $\mathbf{9.2}$ & - \\
  \bottomrule
  \multicolumn{4}{l}{*Does not use any training data.}\\
\end{tabular}
}
\caption{Evaluation of generalization capabilities of supervised models across scenes and domains.}
\label{table:cross_domain_perf}
\end{table}
 
We use three supervised models as our baselines. Namely, we evaluate a two-stream CNN architecture to model spatial and temporal dynamics for saliency-based attention prediction. 
We also add an LSTM-based predictor to the two-stream model to add a saccade model on top of the fixation points provided by the CNN models. 
Finally, we train a simple CNN-LSTM model to predict the gaze position from RGB videos. 
Combined, these three baselines represent the basic architectures used in the gaze prediction task using deep learning models. 
We quantify performance using the Average Angular Error (AAE) and summarize the results in Table~\ref{table:cross_domain_perf}. We denote GTEA Gaze+ as the \textit{same domain} data and GTEA Gaze as the \textit{different domain} data. 
We use the GTEA Gaze+ dataset as the training domain since it has significantly more annotated data than the GTEA Gaze dataset. 

It can be seen that increasing the complexity of the model provides exciting results on scenes from the same domain (i.e., testing within the same dataset) but also suffers a tremendous loss of performance when testing on scenes from different domains. 
The 2-Stream CNN models perform very well on the same domain, achieving an average angular error as low as $5.52$, but also sees a significant drop ($5.83$ and $5.36$, with and without an LSTM predictor, respectively). The simple CNN-LSTM baseline performs reasonably well on the same domain data without any bells and whistles but performs worse than saliency-based models on out-of-domain data. In fact, all three models perform worse than \textit{Center Bias}, which always predicts the center point of the frame as the gaze position. 
We also show the performance of our unsupervised model that does not use \textit{any training data}. Our model performs well across both domains without significant loss in performance. 
\section{Energy-based Surprise Modeling}
In this section, we introduce our energy-based surprise modeling approach for gaze prediction, as illustrated in Figure~\ref{fig:overallArch} and described in Algorithm~\ref{alg:Gaze_pred}. 
We first introduce the necessary background on the pattern theory representation and present the proposed gaze prediction formulation. 

\subsection{Pattern Theory Representation.}
\textbf{Representing Features:} Following Grenander's notations (\cite{grenander1996elements}), the basic building blocks of our representation are atomic components called \textit{generators} ($g_i$). The collection of all possible generators in a given environment is termed the \textit{generator space} ($G_S$). While there can exist various types of generators, as explored in prior pattern theory approaches~\cite{aakur2019generating}, we consider only one type of generator, namely the \textit{feature} generator. 
We define feature generators as features extracted from videos and are used to estimate the gaze at each time step. Each generator represents both appearance- and motion-based features at different spatial locations in each frame of the video.

\textbf{Capturing Local Regularities.} We model temporal and spatial associations among generators through \textit{bonds}. 
Each generator $g_i$ has a fixed number of bonds called the \texttt{arity} of a generator ($w(g_i) \forall g_i \in G_S$). 
These bonds are symbolic representations of the structural and contextual relationships shared between generators. 
Each bond is directed and are differentiated through the direction of information flow as either  \textit{in-bonds} or \textit{out-bonds}. 
Each bond is identified by a unique coordinate and bond value such that the \emph{$j^{th}$} bond of a generator $g_i \in G_S$ is denoted as $\beta_{dir}^{j}(g_i)$, where $dir$ denotes the direction of the bond. 

The \emph{energy} of a bond is used to quantify the strength of the relationship expressed between two generators and is given by the function:
\begin{equation}
b_{struct}(\beta^{\prime}(g_i),\beta^{\prime\prime}(g_j)) = w_s\tanh(\phi(g_i,g_j)). 
\label{SemEnergy}
\end{equation}
where $\beta^{\prime}$ and $\beta^{\prime\prime}$ represent the bonds from the generators $g_i$ and $g_j$, respectively; $\phi(\cdot)$ is the strength of the relationship expressed in the bond; and $w_s$ is a constant used to scale the bond energies. In our framework, $\phi(\cdot)$ is a distance metric and provides a measure of similarity between the features expressed through their respective generators. 
Generators combine through their corresponding bonds to form complex structures called \textit{configurations}. Each configuration has an underlying graph topology, specified by a connector graph $\sigma \in \Sigma$, where $\Sigma$ is the set of all available connector graphs. $\sigma$ is also called the connection type and is used to define the directed connections between the elements of the configuration. In our work, we restrict $\Sigma$ to a \textit{lattice} configuration, as illustrated in Figure~\ref{fig:overallArch}, with bonds extending spatially and aggregated temporally.

\subsection{Local Configuration Proposal}\label{sec:featConfig}\label{sec:localProp}
The first step in the framework is the construction of a lattice configuration called the \textit{feature configuration} ($\hat{c}$). 
The lattice configuration is a $N\times N$ grid, with each point (generator) in the configuration representing a possible region of fixation. 
We construct the feature configuration at time $t$ (defined as $\hat{c}(t)$) by dividing the input image into an $N\times N$ grid. 
Each generator is populated by extracting features from each of these grids. 
These features can be appearance-based such as RGB values, motion features such as optical flow~\cite{brox2010large}, or deep learning features~\cite{redmon2016you}. We experiment with both appearance and motion-based features and find that motion-based features (Section ~\ref{sec:results}) provide distracting cues, especially with significant camera motion associated with egocentric videos. Bonds are quantified across generators with the spatial locality, i.e., all neighboring generators are connected through bonds. 
The bond energy is set to $1$ by default. 

\begin{algorithm}
    \SetKwInOut{Input}{Input}
    \SetKwInOut{Output}{Output}

    Gaze Prediction $(I_t, G_s, C_{t-1}, k, p, p_c, c_b)$\;
    $\hat{c}(t) \leftarrow  featureConfig(I_t, G_s)$\\
    $c^{\prime}(t) \leftarrow temporalAggregate(\hat{c}_t, C_{t-1})$\\
    $ t \leftarrow UniformSample(0,1) $\\
        \uIf {$t < p_c$}
        {
            $c(t) \leftarrow c^\prime(t)$
        }
        \uElse{
            $c(t) \leftarrow c_b$ 
        }
    $G_i(t) \leftarrow \{g_1,g_2,\ldots,g_n \in c(t)\}$\\
    $g_{pred(t)} \leftarrow selectGenerator(G_i(t))$\\
    
    \ForEach{$g_i \in c(t)$}{
        \uIf{$E(g_{pred} | c(t)) < E(g_{i} | c(t)) $}{
        $ t \leftarrow UniformSample(0,1) $\\
        
        \uIf {$t < p$}
        {   
            $E(g_i) \leftarrow \frac{E(g_i)}{d(g_{pred}(t), g_{pred}(t-1))}$\\
            $g_{pred}(t) \leftarrow g_i$
        }
        }
    }
    \Return $g_{pred}(t)$
  \caption{Gaze Prediction using Pattern Theory}
  \label{alg:Gaze_pred}
\end{algorithm}
A local proposal for the time steps $t$ is done through a  temporal aggregation of the previous $k$ configurations across time. The aggregation enforces temporal consistency into the prediction and selectively influences the current prediction. Bonds are established across time by computing bonds between generators with a spatial locality and are quantified using the energy function defined in Equation~\ref{SemEnergy}. 
For example, a bond can be established with generator $g_i(t)$ in feature configuration at time $t$ and the configuration at time $t-1$. 
We define $\phi(g_i,g_j)$ to be a metric that combines both appearance-based and motion-based features. We ensure that the function $\phi(g_i,g_j)$ produces nonzero values proportional to the degree of correlation between the features, both spatially and temporally and is formally defined as
\begin{equation}
\phi(g_i, g_j) = min(\alpha, \phi_a(g_i, g_j) + \phi_m(g_i, g_j))
\end{equation}
where $\alpha$ is a modulating factor to ensure that the energy does not explode to infinity, 
$\phi_a(g_i, g_j)$ and $\phi_m(g_i, g_j)$ are appearance-based and motion-based features, respectively; 
$\phi_a(g_i,g_j)$ is used to quantify the appearance-based affinity and is given by 
$\phi_a(g_i,g_j) = 1 - \frac{ \sum_{i=1}^{N}{{\bf g}_i{\bf g}_j} }{ \sqrt{\sum_{i=1}^{N}{({\bf g}_i)^2}} \sqrt{\sum_{i=1}^{N}{({\bf g}_j)^2}} }$, and is used to capture the spatial correlation between the visual features at a given location in the current frame. 
On the other hand, $\phi_m(g_i,g_j)$ is used to quantify the motion-based bond affinity and is given by $\phi_m(g_i,g_j) = \frac{ \sum_{i=1}^{N}{({{\bf g}_i - \bf {g}}_i)({{\bf g}_j - \bf {g}}_j)}}{ N }$. 
Both metrics produce nonzero values proportional to the degree of correlation. Hence, their use in Equation~\ref{SemEnergy} ensures that the bond energy is reflective of the \textit{predictability} of generators at various time steps.

The temporal consistency is enforced through an ordered weighted averaging aggregation operation of configurations across time. Formally, the temporal aggregation is a mapping function $F : c_n \longrightarrow c$ that maps $n$ configurations from previous time steps into a single configuration as a local proposal for time $t$. The function $F$ has an associated set of weights $W = [w_1, w_2, \ldots w_n]$ lying in the unit interval and summing to one. $W$ is used to aggregate configurations across time using the function $F$ given by
\begin{equation}
    F(c_{t-i}, \ldots c_t) = \sum_{i=1}^{k} w_i (c_t \odot c_{t-i})
    \label{eqn:temporalAgg}
\end{equation}
where $\odot$ refers to pairwise aggregation of bonds across configurations $c_t$ and $c_{t-i}$. $W$ is a set of weights used to scale the influence of each configuration from the past on the current prediction. 
$W$ is set to be an exponential decay function given by $a(1-b)^k$, where $b$ is the decay factor and $a$ is the initial value. We set $k$ to video \textit{fps}, $a$ and $b$ are is set to $1$ and $0.95$. 
Hence, the temporally local configuration has a more significant influence on the current configuration proposal compared to temporally distant configurations. 

\subsection{Gaze Prediction}\label{sec:gazePred}
Intuitively, each generator in the configuration corresponds to the predictability of each spatial segment in the image. Hence, the predicted gaze position corresponds to the grid cell with most \textit{surprise}, i.e., the generator with maximum energy. We begin by constructing the initial feature configuration for time $t$ and temporally aggregating it to obtain the initial configuration $c\prime(t)$ as described in Section~\ref{sec:localProp}. 
Algorithm~\ref{alg:Gaze_pred} illustrates the process of finding the generator with maximum energy. $C_{t-1}, I_t, G_s$ refer to the set of past $k$ configurations, the video frame at time $t$, and the generator space, respectively. 

In addition to na\"{i}ve surprise modeling, we introduce additional constraints to more closely model human gaze, such as the choice between \textit{saccade} and \textit{fixation} using two approaches. 
First, we do so by having two acceptor functions (lines $5-8$ and $12-15$). 
In the first function, the local temporal proposal is accepted with a probability of $p_c$ or rejected in favor of a configuration with strong center bias ($c_b$) as the final prediction for time $t$. The role of this acceptor function is to prevent the model from being distracted due to spurious patterns in the input visual stream. 
In the second acceptor function, we allow the model to switch between saccade and fixation modes by not forcing the model to always choose the generator with the highest energy as the gaze position. 
 
Every newly proposed generator is accepted if it passes the test at line 12-14, which is either true for new generators with energy lower than the current generator's or true with a certain probability ($p$) that is proportional to the energy difference between the recent and the old generators.
Second, we scale each generator's energy with a distance function ($d(\cdot)$ in line 15) that quantifies the distance between the previous predicted gaze point and the current generator. This scaling ensures that the model can fixate on a chosen target while allowing for the saccade function to select a different target if required. 
Once the generator ($g_{pred}$) is chosen, the final gaze point ($x_{pred}, y_{pred}$) is computed as
\begin{equation}
(x_{pred}, y_{pred}) = (c_x + 0.5*W_g, c_y + 0.5*H_g)
\end{equation}
where $(c_x, c_y)$ is the offset of the grid from the top left corner of the image; ($W_g,H_g$) are grid width and height respectively. 
\section{Experimental Evaluation}\label{sec:results}
In this section, we describe the experimental setup used to evaluate our approach and present quantitative and qualitative results on both gaze prediction and event segmentation.
\subsection{Data and Evaluation Setup}
\textbf{Data.} We evaluate our approach on the \textbf{GTEA}~\cite{fathi2012learning} and \textbf{GTEA+}~\cite{li2013learning} datasets. The two datasets consist of video sequences on meal preparation tasks. GTEA contains $17$ sequences of tasks from $14$ subjects, with each sequence lasting about $4$ minutes. GTEA+ contains longer sequences of $5$ subjects performing $7$ activities. We use the official splits for both GTEA and GTEA+ as defined in prior works ~\cite{fathi2012learning,li2013learning}. 

\textbf{Evaluation Metrics.} We use \textbf{Average Angular Error} (AAE) ~\cite{riche2013saliency} as our primary evaluation metric, following prior efforts~\cite{itti2006bayesian,fathi2012learning,li2013learning}. AAE is the angular distance between the predicted gaze location and the ground truth. \textbf{Area Under the Curve} (AUC)  measures the area under a curve for true positive versus false-positive rates under various threshold values on saliency maps. AUC is not directly applicable since our prediction is not a saliency map.

\subsection{Baseline Approaches and Ablation}
\textbf{Baselines.} We compare with state-of-the-art unsupervised and supervised approaches. We consider unsupervised saliency models such as 
Graph-Based Visual Saliency (GBVS)~\cite{harel2007graph},  Attention-based Information Maximization (AIM)~\cite{bruce2006saliency}, 
Itti's model ~\cite{itti2000saliency}, 
Adaptive Whitening Saliency (AWS)~\cite{leboran2016dynamic},  Image Signature Saliency (ImSig)~\cite{hou2011image}, 
OBDL~\cite{hossein2015many} and AWS-D~\cite{leboran2016dynamic}. 
We also compare against supervised models (DFG~\cite{zhang2017deep}, Yin~\cite{li2013learning} and SALICON~\cite{jiang2015salicon}, LDTAT~\cite{huang2018predicting}) that leverage annotations and representation learning capabilities of deep neural networks. 

\textbf{Ablation.} We perform ablations of our approach to test the effectiveness of each component. We evaluate the effect of different features by including optical flow~\cite{brox2010large} as input to the model. We remove the prior spatial constraint and term the model \textit{``Ours (Saccade)''}, highlighting its tendency to saccade through the video without fixating.
\subsection{Quantitative Evaluation}
\begin{table}
\centering
\resizebox{0.99\columnwidth}{!}{
\begin{tabular}{|c|c|c|c|}
\hline
\multirow{2}{*}{\textbf{Supervision}} & \multirow{1}{*}{\textbf{Approach}}      & \textbf{GTEA}  & \textbf{GTEA+}\\ 
  &    & (\textbf{AAE})  & (\textbf{AAE})\\ 
  \toprule
    \multirow{10}{*}{None} 
& \textbf{Ours} & \textbf{9.2} & \textbf{11.6} \\
& Ours (Optical Flow) & 10.1 & 13.8 \\
& Ours (Saccade) & 12.5 & 14.6 \\
& AIM & 14.2 & 15.0 \\
& GBVS& 15.3 & 14.7\\
& OBDL& 15.6 & 19.9\\
& AWS& 17.5 & 14.8\\
& AWS-D& 18.2 & 16.0\\
& Itti's model& 18.4 & 19.9 \\
& ImSig& 19.0 & 16.5\\
 \midrule
\multirow{3}{*}{Full} 
& LDTAT & \textbf{7.6} & \textbf{4.0}\\
& Yin & 8.4 & 7.9\\
& DFG & 10.5 & 6.6\\
& SALICON & 16.5 & 15.6\\
\bottomrule
\end{tabular}
}
\caption{Evaluation on GTEA and GTEA+ datasets. We outperform all unsupervised and some supervised baselines.}
\label{table:perf}
\end{table}
We present the results of our experimental evaluation in Table~\ref{table:perf}. We present the Average Angular Error (AAE) for both the GTEA and GTEA+ datasets and group the baseline approaches into two categories - \textit{no supervision} and \textit{full supervision}. Our approach outperforms all unsupervised methods on both datasets. 
The overall performance on GTEA Gaze is lower than that on GTEA Gaze Plus, which could arguably be attributed to the fact that $25\%$ of ground truth gaze measurements are missing. 
Note that the gap between the performance of fully supervised models between the two datasets is high and shows that they suffer from the lack of a large number of training examples, which is significantly lower in GTEA. 

It is interesting to note that our model outperforms saliency based methods, including the closely related graph based visual saliency model and Itti's model. We attribute it to the use of multiple acceptor functions (Section~\ref{sec:gazePred}), which allows us to switch between fixation and saccade modes and hence is not hindered by saliency-based features that do not capture task-dependent attention. 
Note that we significantly outperform SALICON~\cite{jiang2015salicon}, a fully supervised convolutional neural network trained on ground-truth saliency maps on both datasets. We also outperform the GAN-based gaze prediction model DFG~\cite{zhang2017deep} on the GTEA dataset, where the amount of training data is limited. 
We also offer competitive performance to Yin \emph{et al.}~\cite{li2013learning}, who use auxiliary information in the form of visual cues such as hands, objects of interest, and faces. This gap is particularly narrowed when considering domains with lower data (GTEA Gaze) and the lack of generalization (Section~\ref{sec:motivation}) shown by common deep learning approaches to domains outside their training environment. It should be noted that our approach runs at 45 fps in a \textit{single threaded application}, run on an AMD ThreadRipper CPU, a significantly accelerated method compared to deep learning approaches and offers a way forward for resource-constrained prediction.
\begin{table}
\caption{Evaluation on the temporal video segmentation task on the GTEA Gaze datasets.}
\label{table:seg_perf}
\centering
\resizebox{0.99\columnwidth}{!}{
\begin{tabular}{|c|c|c|}
\hline
\multirow{1}{*}{\textbf{Supervision}} & \multirow{1}{*}{\textbf{Approach}}      & \textbf{Accuracy}\\ 
  \toprule
    \multirow{3}{*}{Full} & Spatial-CNN (S-CNN) & $54.1$\\
    & Bi-LSTM~\cite{lea2016temporal} & $55.5$\\
    & EgoNet~\cite{singh2016first} & $57.6$\\
    & Dilated TCN~\cite{lea2016temporal} & $58.3$\\
    & TCN~\cite{lea2016temporal} & $\mathbf{64.1}$\\
 \midrule
\multirow{2}{*}{None} & Ours w/ 3D-CNN features & $\mathbf{35.17}$\\
& Ours (Streaming Eval.) & $57.92$\\

\bottomrule
\end{tabular}
}
\end{table}

\subsection{Video Event Segmentation}
To highlight our model's ability to understand video semantics and its subsequent ability to predict gaze locations, we adapt the framework to perform event segmentation in streaming videos. Instead of localizing specific generators, we monitor the \textit{global surprise} by considering the energy of the entire configuration $c(t)$. 
The energy of a configuration $c$ is the sum of the bond energies (Equation \ref{SemEnergy}) in a configuration and is given by 

\begin{equation}
E(c) = -\sum_{ (\beta^{\prime},\beta^{\prime\prime})\in c}{b_{struct}(\beta^{\prime}(g_i),\beta^{\prime\prime}(g_j))}
\label{energy}
\end{equation}
where a lower energy indicates that the generators are closely associated with each other. Hence, a higher energy suggests that the \textit{surprise} faced by the framework is higher. The configuration (frame) with the highest energy is considered to hold the highest \textit{surprise} and is selected as an event boundary. We use the error gating-based segmentation used in ~\cite{aakur2019perceptual}. We set the error threshold to be $2.5$ as opposed to $1.5$, and the energy of the configuration as the perceptual quality metric for event segmentation. 
We evaluate the performance of the approach on the GTEA dataset and quantify its performance using accuracy. We use a 3D-CNN~\cite{ji20123d} network pre-trained on UCF-101~\cite{soomro2012ucf101} to extract features. 
We use k-means clustering to assign each segment to a cluster label. 
We then align the prediction and ground-truth using the Hungarian method before computing accuracy, following prior works~\cite{lea2016temporal}. 
As can be seen from Table~\ref{table:seg_perf}, our \textit{unsupervised} approach achieves a temporal segmentation accuracy of \textbf{$35.17\%$}. We compare against several state-of-the-art \textit{supervised} deep learning approaches such as Spatial-CNN, Bi-LSTM, and Temporal Convolutional Networks (TCN)~\cite{lea2016temporal}, which achieve accuracy of $54.1\%$, $55.5\%$ and $64.1$, respectively. Note that the class-agnostic accuracy, i.e., when \textit{evaluated in a streaming manner per video}, we obtain a segmentation accuracy of \textbf{$57.92\%$}, which suggests that the actual segmentation is robust. 

\subsection{Qualitative Evaluation}
\begin{figure*}
\centering
\begin{tabular}{ccccc}
\includegraphics[width=0.17\textwidth]{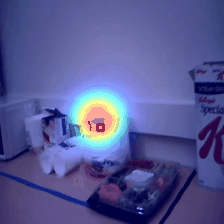} &
  \includegraphics[width=0.17\textwidth]{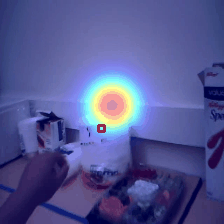} & \includegraphics[width=0.17\textwidth]{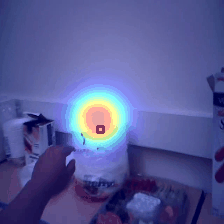} & 
  \includegraphics[width=0.17\textwidth]{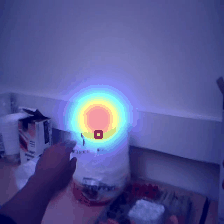} & \includegraphics[width=0.17\textwidth]{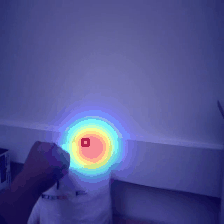} \\
\multicolumn{5}{c}{(a)}\\
\includegraphics[width=0.17\textwidth]{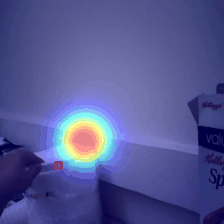} &
  \includegraphics[width=0.17\textwidth]{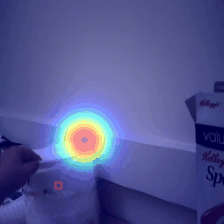} & \includegraphics[width=0.17\textwidth]{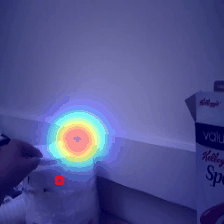} & 
  \includegraphics[width=0.17\textwidth]{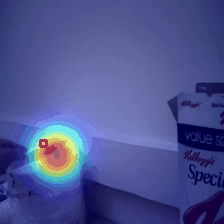} & \includegraphics[width=0.17\textwidth]{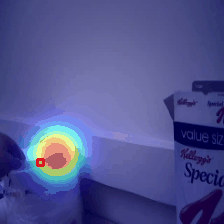} \\ 
\multicolumn{5}{c}{(b)}\\
\includegraphics[width=0.17\textwidth]{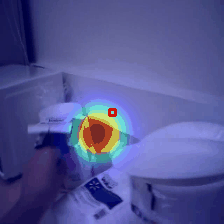} &
  \includegraphics[width=0.17\textwidth]{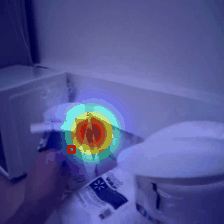} & \includegraphics[width=0.17\textwidth]{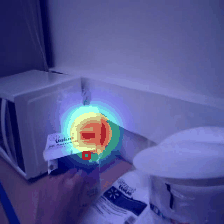} & 
  \includegraphics[width=0.17\textwidth]{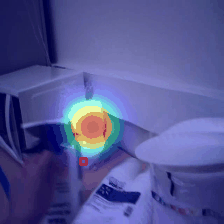} & \includegraphics[width=0.17\textwidth]{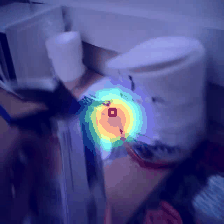} \\ 
\multicolumn{5}{c}{(c)}\\
\end{tabular}
\caption{\textbf{Qualitative Examples}: We present the predictions made by our model for three different sequences across different tasks and domains. The predictions are shown as a gaussian map and the ground truth is highlighted in red.}
\label{fig:qual_res}
\end{figure*}

We present some qualitative visualizations in Figure~\ref{fig:qual_res}. We illustrate three cases from our model's predictions. The ground-truth gaze position is shown in red, and our predictions are shown as a heatmap. It is interesting to note (particularly highlighted in Figure~\ref{fig:qual_res}(b)) is the tendency of our model to fixate on highly relevant regions and subsequent objects, even though the ground-truth gaze positions vary. The predicted gaze then quickly adapts to the newer position and starts fixating on the relevant object. This characteristic suggests that despite purely bottom-up visual processing, the notion of ``\textit{surprise}'' has an underlying task-bound nature. 
We attribute this to the two acceptor functions (lines $5-8$ and $12-15$) defined in Algorithm~\ref{alg:Gaze_pred}). The former allows us to enforce some spatial bias into the model and force the gaze prediction back to the center of the image (the effect is highlighted in Figure~\ref{fig:qual_res}(a)). The latter helps fixate on certain objects for a longer duration and help handle clutter and occlusion for deriving task-specific object affordances, \textit{without any specific training objectives and data}. 

\section{Conclusion}
In this work, we analyze the generalization ability of supervised deep learning models to different domains and scenes for the egocentric gaze prediction task and find that their performance suffers when presented with out-of-domain data. 
To break the increasing dependence on training data, we present one of the first approaches to a unified framework for tackling  \textit{unsupervised} gaze prediction and temporal segmentation, based on energy-based surprise modeling. Using a novel formulation, we demonstrate that pattern theory can be used to predict gaze locations in egocentric videos. Our pattern theory representation also forms the basis for unsupervised temporal video segmentation. 
Through extensive experiments, we demonstrate that we obtain state-of-the-art performance on the unsupervised gaze prediction task and provide competitive performance on the unsupervised temporal segmentation task on egocentric videos. 
\section{Acknowledgement}
This research was supported in part by the US National Science Foundation grant IIS 1955230.

\bibliographystyle{apalike}
{\small
\bibliography{egbib}}

\end{document}